\documentclass{article}

\usepackage{arxiv}

\usepackage[utf8]{inputenc} 
\usepackage[T1]{fontenc}    
\usepackage{hyperref}       
\usepackage{url}            
\usepackage{booktabs}       
\usepackage{amsfonts}       
\usepackage{nicefrac}       
\usepackage{microtype}      
\usepackage{lipsum}
\usepackage{graphicx}
\graphicspath{ {./figs/} }

\usepackage{amsmath}
\usepackage{caption}
\usepackage{subcaption}
\usepackage{xcolor}
\usepackage{tabularx}

\title{Development of deep transformer-based models for long-term prediction of transient production of oil wells}

\author{
    Ildar R. Abdrakhmanov \\
    \texttt{Ildar.Abdrakhmanov@skoltech.ru} \\
    \And
    Evgenii A. Kanin \\
    \texttt{e.kanin@skoltech.ru} \\
    \And
    Sergei A. Boronin \\
    \texttt{s.boronin@skoltech.ru} \\
    \And
    Evgeny V. Burnaev \\
    \texttt{e.burnaev@skoltech.ru} \\
    \And
    Andrei A. Osiptsov \\
    \texttt{a.osiptsov@skoltech.ru} \\
    \AND
    \\
    Skolkovo Institute of Science and Technology (Skoltech) \\
    Bolshoy Boulevard 30, bld. 1, Moscow, Russia, 121205

}

\begin{document}
\maketitle
\begin{abstract}
We propose a novel approach to data-driven modeling of a transient production of oil wells. We apply the transformer-based neural networks trained on the multivariate time series composed of various parameters of oil wells measured during their exploitation. By tuning the machine learning models for a single well (ignoring the effect of neighboring wells) on the open-source field datasets, we demonstrate that transformer outperforms recurrent neural networks with LSTM/GRU cells in the forecasting of the bottomhole pressure dynamics. We apply the transfer learning procedure to the transformer-based surrogate model, which includes the initial training on the dataset from a certain well and additional tuning of the model’s weights on the dataset from a target well. Transfer learning approach helps to improve the prediction capability of the model. Next, we generalize the single-well model based on the transformer architecture for multiple wells to simulate complex transient oilfield-level patterns. In other words, we create the global model which deals with the dataset, comprised of the production history from multiple wells, and allows for capturing the well interference resulting in more accurate prediction of the bottomhole pressure or flow rate evolutions for each well under consideration. The developed instruments for a single-well and oilfield-scale modelling can be used to optimize the production process by selecting the operating regime and submersible equipment to increase the hydrocarbon recovery. In addition, the models can be helpful to perform well-testing avoiding costly shut-in operations.
\end{abstract}


\section{Introduction}
Large datasets are usually accumulated during an oilfield life cycle. Typically, they are stored as time series and composed of various parameters measured at producing wells by pressure gauges, temperature sensors, and flow meters. This is a rich source of information that can be used to develop models to simulate well performance and predict the production process. Finding the relationship between bottomhole pressure and flow rate at the surface is a primary task in the petroleum industry which has several technological applications. Its solution is usually carried out by constructing mathematical models based on conservation laws in fluid mechanics supplemented by closure relations obtained from the analysis of experimental data. The models can be either semi-analytical \cite{ozkan1991new_1, ozkan1991new_2} or fully numerical obtained by simulators accounting for the multiphase flow in porous media, PVT properties of the phases, and other physical phenomena accompanying hydrocarbon recovery (e.g., Eclipse, TNavigator). In both cases, additional information is required related to a reservoir, fluids properties as well as geometries of the completion and formation, which can be extracted from the interpretation of well-testing, logging, and seismic survey data. Hydrodynamic simulators imply specific competency to petroleum engineers, and the computation process is usually time-consuming, especially when massive parametric studies are carried out.

In this work, we develop a novel data-driven model based on the deep learning algorithms, which allow fast prediction of time-dependent well parameters such as bottomhole pressure or flow rate. It can be an alternative to relatively slow running numerical simulators based on mechanistic models. The developed instrument is not a replacement for the traditional fluid mechanics-based modeling, but it potentially expands the capability of petroleum engineers for planning the development of the oil and gas fields. 

There are many literature sources devoted to data-driven modeling of the transient production of oil wells. For example, in the paper \cite{li2019deep}, the authors used time series with a moderate duration to predict the bottomhole pressure dynamics. It was obtained that long- and short-term time-series network (LSTNet) \cite{lai2018modeling} outperforms the recurrent neural networks (RNNs) trained on the field data for a single well in terms of prediction accuracy. The authors analyzed different feature combinations and highlighted the optimal set providing the maximum quality of predictions. In the study \cite{alakeely2020simulating}, the authors compared the predictive capabilities of convolutional (CNN) and recurrent (RNN) neural networks to simulate the reservoir behavior. The authors demonstrated that both networks could capture the interference between multiple wells at the same reservoir. RNN with the long-short-term memory (LSTM) cell \cite{hochreiter1997long} was applied in the paper \cite{madasu2018deep} to predict the wellhead pressure during a fracturing treatment, and the input parameters of the network are the surface characteristics of the well. In the papers discussed above, the machine learning models utilize either recurrent or convolutional neural networks. However, RNNs have exploding and vanishing gradient problems, while the convolutional filters in CNNs have certain limitations. These issues restrict the applicability of RNNs and CNNs to model complex long-term dependencies in sequential data series. Auto-regression (AR) methods are another approach to sequence modeling and time-series predictions. The commonly known autoregressive moving average (ARMA) and autoregressive integrated moving average (ARIMA) are adapted to model trends, patterns, and seasonality. As a result, they have limitations in modeling transient production processes with several extreme values in the well’s measurement data. The authors of the paper \cite{tian2019applying} reported another approach for oil well performance modeling. The researchers built the machine learning model based on linear regression to predict the bottomhole pressure using the nonlinear flow-rate features. To reflect the physical properties of the pressure response as a function of flow rate and time, the authors extracted several features from the time-series data. The feature-based machine learning methods such as linear regression require manual feature engineering, leading to the loss of important information contained in raw data. In the literature, different authors also applied the widespread machine learning techniques such as the Gradient Boosting algorithm and/or Artificial Neural Network to predict the bottomhole pressure \cite{kanin2019predictive, baryshnikov2020adaptation} of a well and the flow rate after the hydraulic fracturing treatment \cite{morozov2020data, erofeev2021ai, duplyakov2022data}.

In the present paper, we apply a novel deep learning algorithm called transformer \cite{vaswani2017attention} to build surrogate models for simulations of oil well performance. Transformer architecture was initially developed for natural language processing (NLP) problems. However, in recent years, researchers adapted transformers for time-series forecasting. In the paper \cite{wu2020deep}, the transformer was used for influenza-like illness prediction. In the article \cite{lim2019temporal}, the authors introduced the temporal fusion transformers for interpretable multi-horizon time-series forecasting. The main idea is to learn temporal relations at different scales, recurrent layers are used for local processing and self-attention layers are responsible for long-term processing. Several modifications of the original transformer were proposed in the work \cite{li2019enhancing} to improve time-series modeling. The authors enhanced the locality of the model with convolutional self-attention and addressed the memory bottleneck problem with LogSparse self-attention layers. In contrast to recurrent neural networks, the transformer does not process data sequentially. Instead, it handles the entire sequence via a multi-head self-attention mechanism. Self-attention enables the transformer to learn data representations efficiently relating to different positions in the input sequence. The transformer-based models have a large potential for more accurate predictions made on the basis of unsteady and noisy data as compared to recurrent neural networks. Moreover, the transformer is much more computationally efficient than RNN since its training can be performed using graphics processing units (GPU).  

Transformer network allows for applying the transfer learning technique relied on the following idea. First, we train the model to predict the target parameter, for example, bottomhole pressure, using the production data from a certain well. Next, we use the tuned weights to initialize the training procedure on the data from a target well. In other words, we carry out the fine-tuning based on the pre-trained model. The described technique allows the model to transfer knowledge from one time-series forecasting problem to another leading to acceleration of the training process and improvement of the model prediction capability. The recurrent networks are not well suited to transfer learning since the hidden state of recurrent cells is commonly not transferred.  

The main aim of the current work is to create an instrument based on transformer architecture for modeling the transient production of an oil well. To the best of our knowledge, the application of transformer networks for the considered problem has never been reported in open literature. In the paper, we propose two models allowing one to simulate the production process of either a single well or several wells (global model). The latter model is trained on data from multiple wells and allows for accounting for interference between wells leading to improvement of accuracy in estimations for each of the wells. We demonstrate that the transformer can learn complex variations in production data and dependencies between time-dependent characteristics of oil wells.

The paper is organized as follows. Section \ref{sec:problem_formulation} outlines the problem formulation. Section \ref{sec:modeling_approach} describes the modeling approach and the field dataset utilized for the training of machine learning models. Section \ref{sec:dl_models} revisits the deep learning algorithms considered in the current work, recurrent neural networks and transformer, to model the transient well performance; the single-well model is improved and expanded via applying the transfer learning and building the global model. Finally, Section \ref{sec:results} presents the obtained results and their discussion including the comparison of the predictive capability of transformer and recurrent neural networks.

\section{Problem formulation}
\label{sec:problem_formulation}
We consider a transient production of an oil well with arbitrary completion. In most cases, various parameters of the well are measured during its exploitation including bottomhole/wellhead pressure and temperature, flow rates for each phase, choke size, parameters of electric centrifugal pump (if available). One of the most important tasks of well performance modeling is to determine the functional relationship between the bottomhole pressure and surface flow rate. The problem can be solved using hydrodynamic simulators based on mathematical models formulated using conservation laws in fluid mechanics. The simulations allow for taking into account different physical phenomena accompanying well production but require values of several additional reservoir and fluids properties typically not measured in the field. Also, the calculations are time-consuming so that they are intended to be run at high-performance computers.

Here, we propose another approach to find the dependence between bottomhole pressure and flow rate. It is based on deep learning algorithms trained on available field data and allows for obtaining quick predictions of bottomhole pressure or flow rate. The data-driven coupled model of a well and reservoir can be used to plan and optimize the production process. Also, the developed model can be used to estimate well and formation properties using a constant flow rate response of a well, which is analog to well-test avoiding expensive shut-ins.     

\section{Modeling approach and dataset}
\label{sec:modeling_approach}
In the current section we describe the developed data-driven model to predict the bottomhole pressure of a single well. The considered problem is an example of a supervised machine learning task. We assume that the field dataset is composed of daily records of the dynamic parameters of a well (in general, the time intervals between the data points may be heterogeneous). We train machine learning algorithms to predict the average value of the bottomhole pressure $\hat{p}_t$ during the next day, and the input data is the sequence of vectors containing the values of governing parameters (flow rates of oil, gas, and water; temperature at the bottomhole and wellhead, and others) at the current time moment t and at previous days $(x_{t-N+1},\ldots,x_{t-1},x_t)$. The surrogate model processes a matrix of dimension $N \times M$, where $N-1$ is the number of days preceding the considered time moment, and $M$ is the number of features. The forecast horizon of the models can be larger than one day, and it is required to use the sequence-to-sequence (Seq2Seq) models with encoder-decoder architecture. However, in the present work, both transformer and recurrent neural networks are trained to predict the one-day value of the target parameter. If the data-driven model to estimate of the flow rate is required (liquid $\hat{Q}_t^l$  or for each phase ${\hat{Q}}_t^o,{\hat{Q}}_t^g,{\hat{Q}}_t^w$), it is formulated in a similar way with the exception that input parameters include the bottomhole pressure dynamics.  

For test calculations we use the data from Volve oilfield in North Sea published by Norwegian company Equinor for research purposes \cite{equinor2018volve}. The data contains the daily measurements of various characteristics of an oil well, and in this work, we build models for two wells operated from April 2014 to April 2016. 

First, we preprocess the data since measurements contain missing values, linear interpolation is applied to fill the gaps. The resulting time series for two oil wells are shown in Appendix. Next, we perform the feature-wise min-max normalization of the entire dataset using the maximum and minimum values of the features determined from the training data as follows:
\begin{equation}
    x_{tj}=\frac{x_{tj}-\min{x_{tj}}}{\max{x_{tj}}-\min{x_{tj}}},
    \label{eq:scaling}
\end{equation}
where $j$ is a feature index. Data normalization is a standard technique in machine learning, especially when neural networks are applied. All input features have different scales, and, therefore, they are rescaled to the range from 0 to 1.

\begin{figure}[h]
   \centering
   \includegraphics[width=0.8\textwidth]{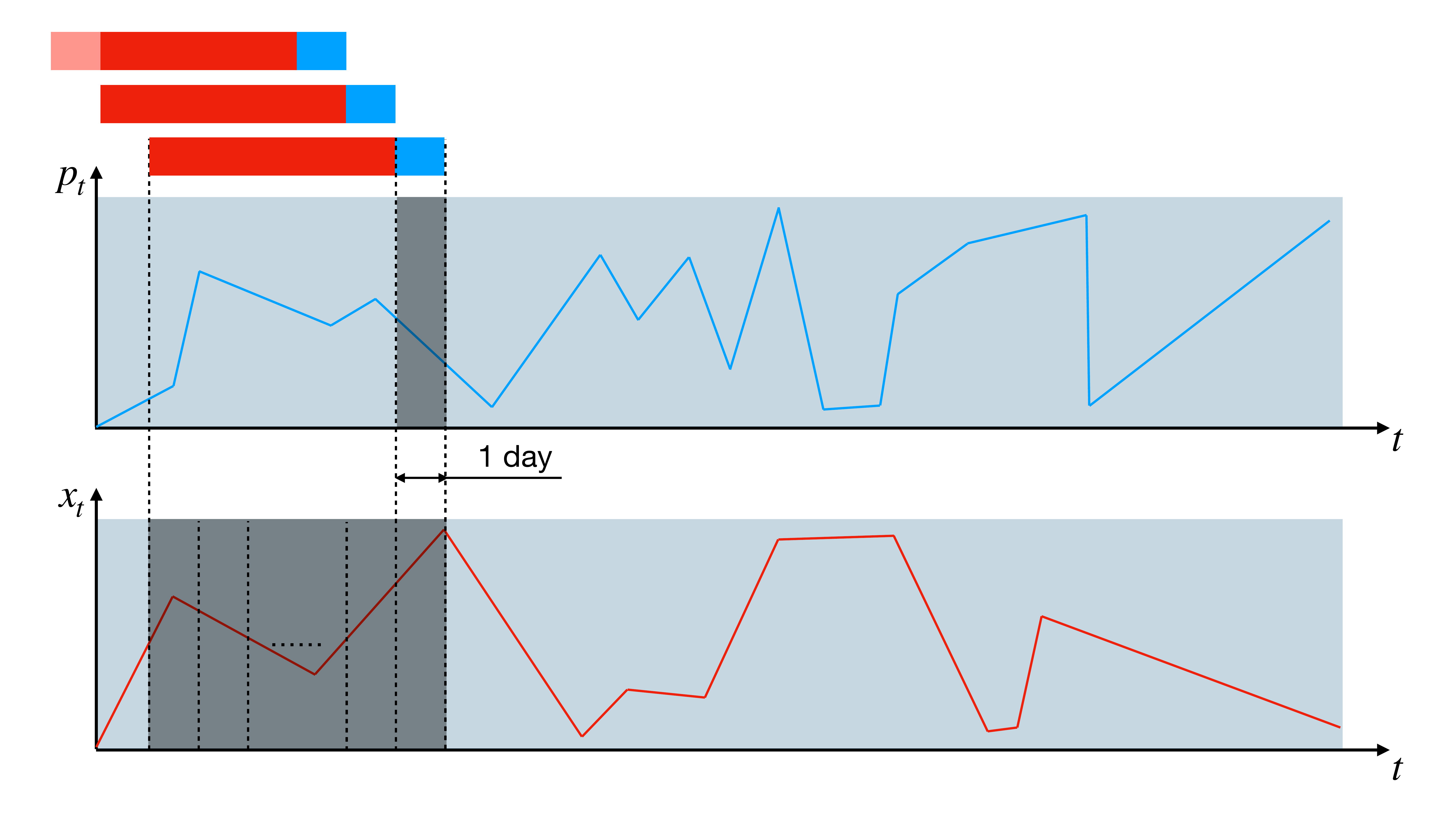}
   \caption{Sliding window technique for cutting the initial time series into fixed-length samples used afterward to train and test machine learning algorithms. Each data sample consists of the time interval (red color) based on which the value of the bottom-hole pressure is predicted for the next day (blue color). The light red color is used for the time intervals lying outside the original production period. The upper graph shows an example of the downhole pressure behavior, while the lower one shows one of the time-dependent governing parameters, e.g., oil flow rate.}
   \label{fig:sliding_window}
\end{figure}

To produce the labeled samples for training and testing of the data-driven models based on deep learning algorithms, we slice the original time series into samples of length $N$ using the sliding window technique. The schematic picture of this procedure is shown in Figure \ref{fig:sliding_window}, where blue color denotes the days for which we predict the bottomhole pressure values, and red color represents the time intervals based on which the prediction is carried out. Several first samples have an incomplete history so that the corresponding segments are padded with zeros (pastel red in Figure \ref{fig:sliding_window}).

Finally, we divide the dataset into three parts: train (70\%), validation (15\%), and test (15\%). One should mention that the validation set follows the training set, while the test set follows the validation set.

\section{Deep learning models}
\label{sec:dl_models}

\subsection{Recurrent neural network}

\begin{figure}[h]
   \centering
   \includegraphics[width=0.8\textwidth]{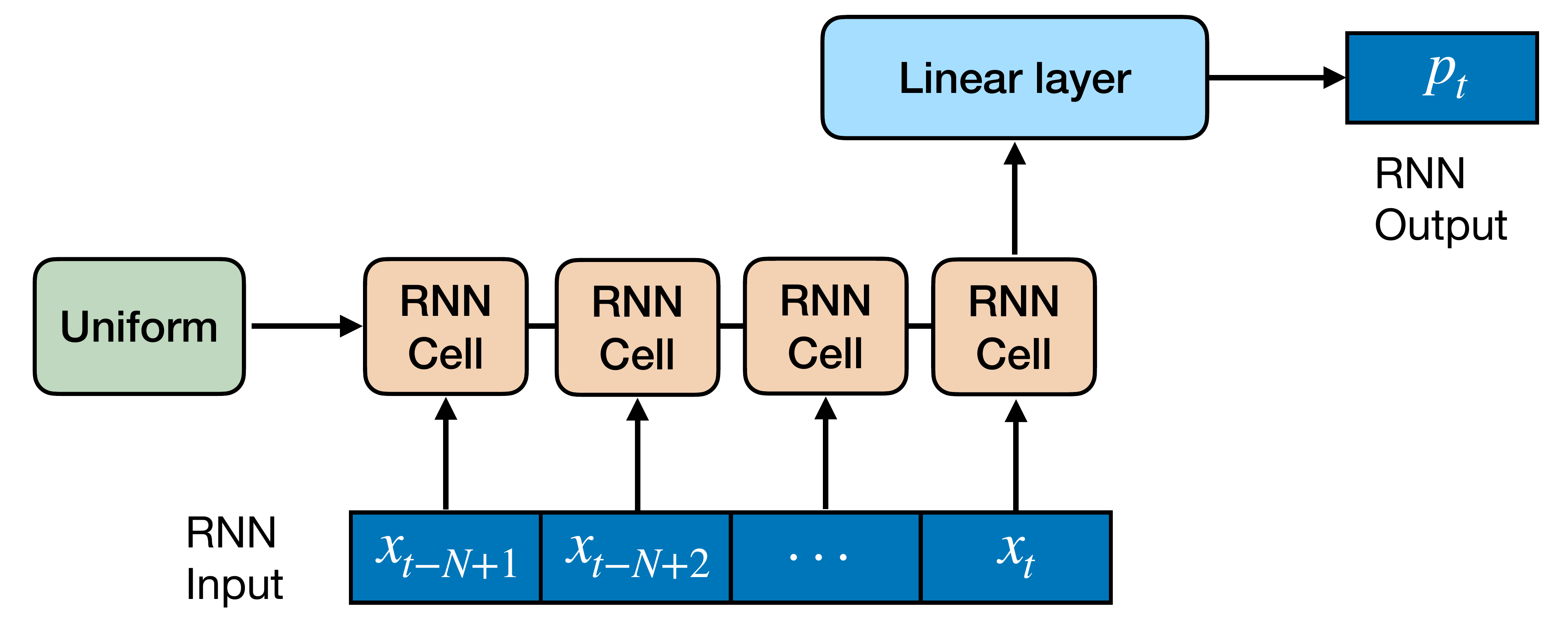}
   \caption{The architecture of many-to-one RNN which converts the input matrix $X_t=\{x_{t-N+1}, \ldots, x_t\}$ with the values of the governing parameters into the bottomhole pressure value $p_t$.}
   \label{fig:rnn_arch}
\end{figure}

The structure of the recurrent neural network (RNN) is shown in Figure \ref{fig:rnn_arch}. 

RNN converts the input matrix into the number, which corresponds to many-to-one architecture. RNN consists of identical sequentially connected blocks (also known as RNN cells), and their number is equal to the length of each data sample (in our case, $N$). Let us consider the simplest RNN cell. It processes the vector of governing parameters $x_i$  and previous hidden state $h_{i-1}$. By applying a certain functional dependence, the RNN cell outputs a new hidden state $h_i$. The last hidden state vector $h_t$  is transformed into the bottomhole pressure value $y_t \equiv p_t$  by a linear layer. We summarize the described transformations by following expressions:
\begin{flalign}
& h_i=\sigma\left(W_x x_i + W_h h_{i-1}+b_h \right), \nonumber \\
& y_t=W_y h_t+b_y,
\label{eq:rnn_equations}
\end{flalign}
where $W_x,W_h,b_h,b_y$ are the matrices and columns with tuned weights. During the training of RNN with the simplest cell, one can observe the gradient vanishing and exploding problems, especially when the input sequence is sufficiently long.

Long-short term memory (LSTM) \cite{hochreiter1997long} and gated recurrent unit (GRU) \cite{cho2014learning} networks were designed to address the vanishing gradient problem via the gates mechanism. The following set of equations describes the LSTM network:
\begin{flalign}
&f_i=\sigma\left(W_f\left[h_{i-1},x_i\right]+b_f\right), \nonumber \\
&i_i=\sigma\left(W_i\left[h_{i-1},x_i\right]+b_i\right), \nonumber \\
&{\hat{c}}_i=\tanh{\left(W_c\left[h_{i-1},x_i\right]+b_c\right)}, \nonumber \\
&c_i=f_i\ast c_{i-1}+i_i\ast{\hat{c}}_i, \nonumber \\
&h_i={\hat{c}}_i\ast\tanh{\left(c_i\right)}, \nonumber \\
& y_t=W_y h_t+b_y,
\end{flalign}
where $f_i$ forgets gate's activation vector,  $i_i$ updates gate's activation vector, ${\hat{c}}_i$ is the cell input activation vector, $c_i$ is the cell state activation vector, $\ast$ is the element-wise product, $W_f,W_i,W_c,W_y,b_f,b_i,b_c,b_y$ are matrices and columns with weights learned during the training.

GRU network has a smaller number of tuned parameters. Only two gates control the information flow in the GRU cells. GRU network is faster and more lightweight than LSTM. The following equations describe the GRU cell:
\begin{flalign}
&z_i=\sigma\left(W_z\left[h_{i-1},x_i\right]+b_z\right), \nonumber \\
&r_i=\sigma\left(W_r\left[h_{i-1},x_i\right]+b_r\right), \nonumber \\
&{\hat{h}}_i=\tanh{\left(W_h\left[r_i\ast h_{i-1},x_i\right]+b_h\right)}, \nonumber \\
&h_i=\left(1-z_i\right)\ast h_{i-1}+z_i\ast{\hat{h}}_i, \nonumber \\
&y_t=W_y h_t+b_y,
\end{flalign}
where $z_i$ updates gate's activation vector, $r_i$ resets gate's activation vector, ${\hat{h}}_i$ is the candidate activation vector, $W_z,W_r,W_h,b_z,b_r,b_h$ are the matrices and bias vectors with parameters learned during training. 

Despite both LSTM and GRU were developed to solve the vanishing gradient problem, the issue is not resolved completely. Exploding gradient problem is tackled with the gradient clipping technique used to rescale the gradient vectors whenever their norm exceeds a certain threshold. We average the gradient norms accumulated over several training epochs to determine the threshold value in our work.

\subsection{Transformer}
\begin{figure}[htbp]
   \centering
   \includegraphics[width=0.8\textwidth]{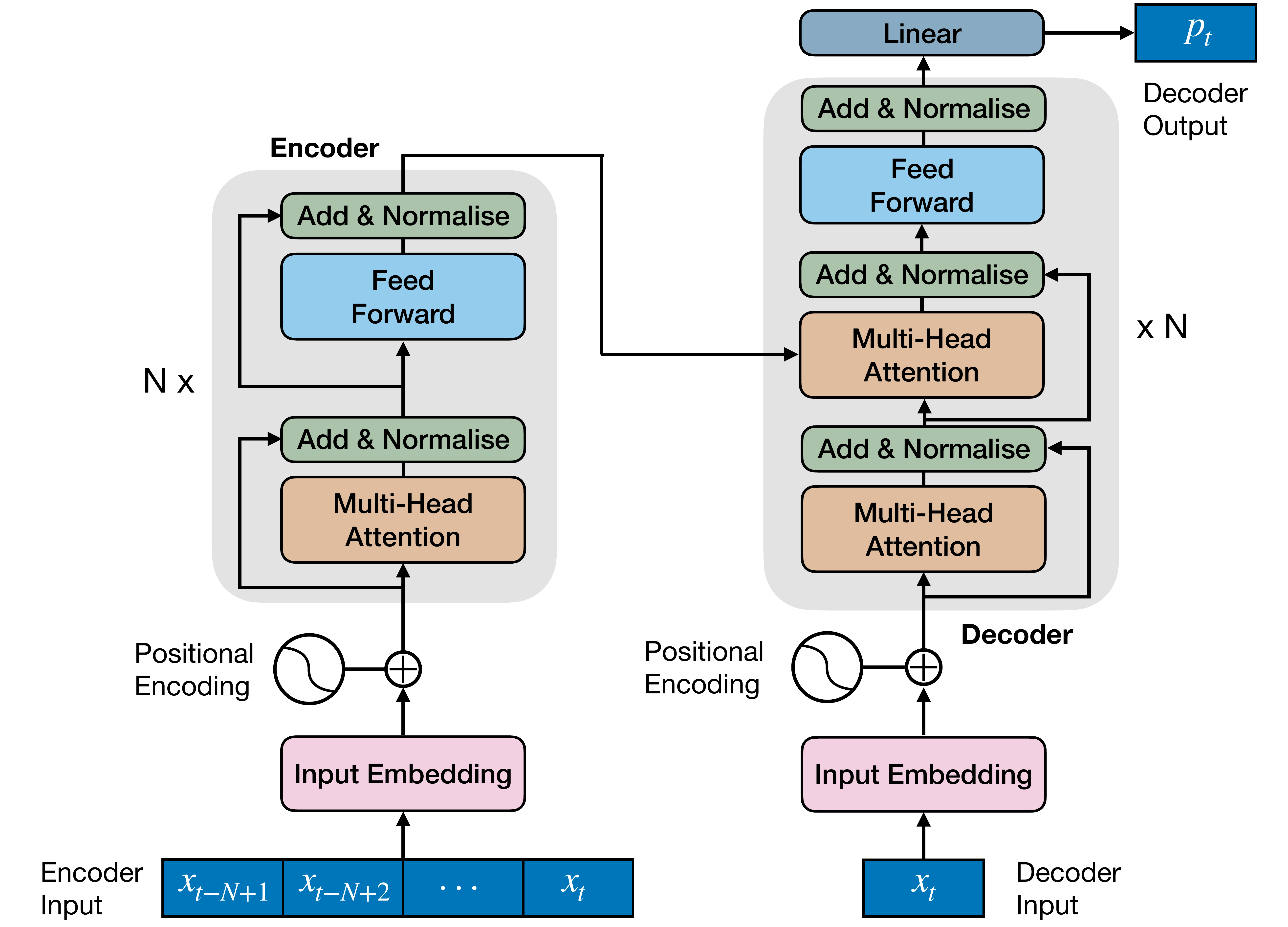}
   \caption{Architecture of transformer-based model: encoder (on the left) and decoder (on the right) internal structures.}
   \label{fig:trans_arch}
\end{figure}

We consider transformer architecture (Figure \ref{fig:trans_arch}) proposed in the original paper \cite{vaswani2017attention} and introduce several modifications to adapt the model for predictions of time-series.
The model consists of encoder and decoder parts. In the former part, the input matrix of dimension $\left[N\times M\right]$ is mapped to the tensor of shape $\left[N\times d_{model}\right]$ through a fully connected layer (input embedding in Figure \ref{fig:trans_arch}). To encode the sequential information in the form of time series, we apply the fixed sinusoidal positional encoding (positional encoding in Figure \ref{fig:trans_arch}). The resulting tensor is passed to the encoder part of the network represented by several encoding layers. Each layer consists of a multi-head self-attention mechanism followed by a fully connected feed-forward network. Similar to the original architecture of the transformer, we use residual connections followed by layer normalization \cite{vaswani2017attention}. The decoder input is the last data point of the encoder input, the vector with the values of the governing parameters corresponding to the time instant for which we perform the prediction. The model predicts the target value at the time instant corresponding to the last data point in the input sequence. The internal algorithm of the decoder resembles that of the encoder. The difference is in the transformation of the decoder output: it is mapped into the value of the target parameter through a linear layer. The number of samples in the dataset is not large compared to the typical NLP tasks. Using the same hyperparameters of the transformer as in the original architecture \cite{vaswani2017attention}, we obtain the model overfitting. To adapt the single-well pressure prediction model, we reduce the number of learned parameters by decreasing the model dimension, number of encoder and decoder layers, number of heads, and dimensions of the feed-forward networks. Note that these numbers in the global model describing multiple wells (Section \ref{sec:transfer_learning}) should be increased due to using of a larger dataset for model training.

\subsection{Transfer learning approach}
Transfer learning is a technique in which the tuned weights of the surrogate model are transferred from one machine learning problem to another. Instead of training the model from scratch, it is initialized with the pre-trained weights. Transfer learning facilitates training and almost always improves the predictive capability of the machine learning model. The hidden state of the recurrent networks is commonly not transferred in stateless RNN models. We utilize stateless RNNs whose hidden state is uniformly initialized before each forward pass. Therefore, we cannot apply the transfer learning technique for such RNN-based models. However, the transformer-based model is suitable for this procedure. We can train the deep learning model on data from one well and then use the pre-trained model to fine-tune it on the data from a different (target) well. First, we get rid of the linear output layer of the transformer and add two new linear layers with ReLU non-linearity between them. The weights of the newly added linear layers are randomly initialized. Large gradients can propagate through the pre-trained transformer model and wreck the pre-trained weights. To avoid this, we run a single warm-up epoch to let the Adam \cite{kingma2014adam} optimizer accumulate the gradient statistics. Then, the whole model is trained with a ten times smaller learning rate as in training from scratch. Correctly performed fine-tuning enhances the quality of the prediction, and we confirm this statement by the results of the numerical experiments shown below.  
\subsection{Global model}
It can be sufficient to train the surrogate model on the measurements from the considered well to simulate the transient production process. However, if we consider a certain well surrounded by neighboring producing/injection wells, any alteration of their operating mode leads to a significant impact on time-dependent parameters of the well. We develop a global model trained on the dataset composed of the measurements from multiple wells. The wells identifiers are encoded into the one-hot vectors, which we include into input features of the model. These new features allow the model to recognize the well, from which the data sample is processed. The data comes into the model in the mini-batches consisting of the data samples belonging to multiple wells. Similar to the single-well model, we utilize the sliding window technique to create data samples (Figure \ref{fig:sliding_window}). The window cuts the fixed time interval from the time series for each well. By incorporating the information from multiple wells, the created model allows capturing the patterns running at the field development cell and/or the entire oilfield scale (if the number of oil wells in the dataset is large) due to an interference between neighboring wells. The global model predicts the values of a time-dependent target parameter (for example, bottomhole pressure) for each considered well more accurately as compared to the single-well model, and we support the statement by the numerical experiments.

\section{Results and discussion}
\label{sec:results}
We conducted a series of numerical experiments using the data for two wells (15/9-F-1C and 15/9-F-15D) from the field dataset described in Section \ref{sec:modeling_approach}. 

\subsection{Single-well model}
\begin{figure}[h]
   \centering
   \includegraphics[width=0.8\textwidth]{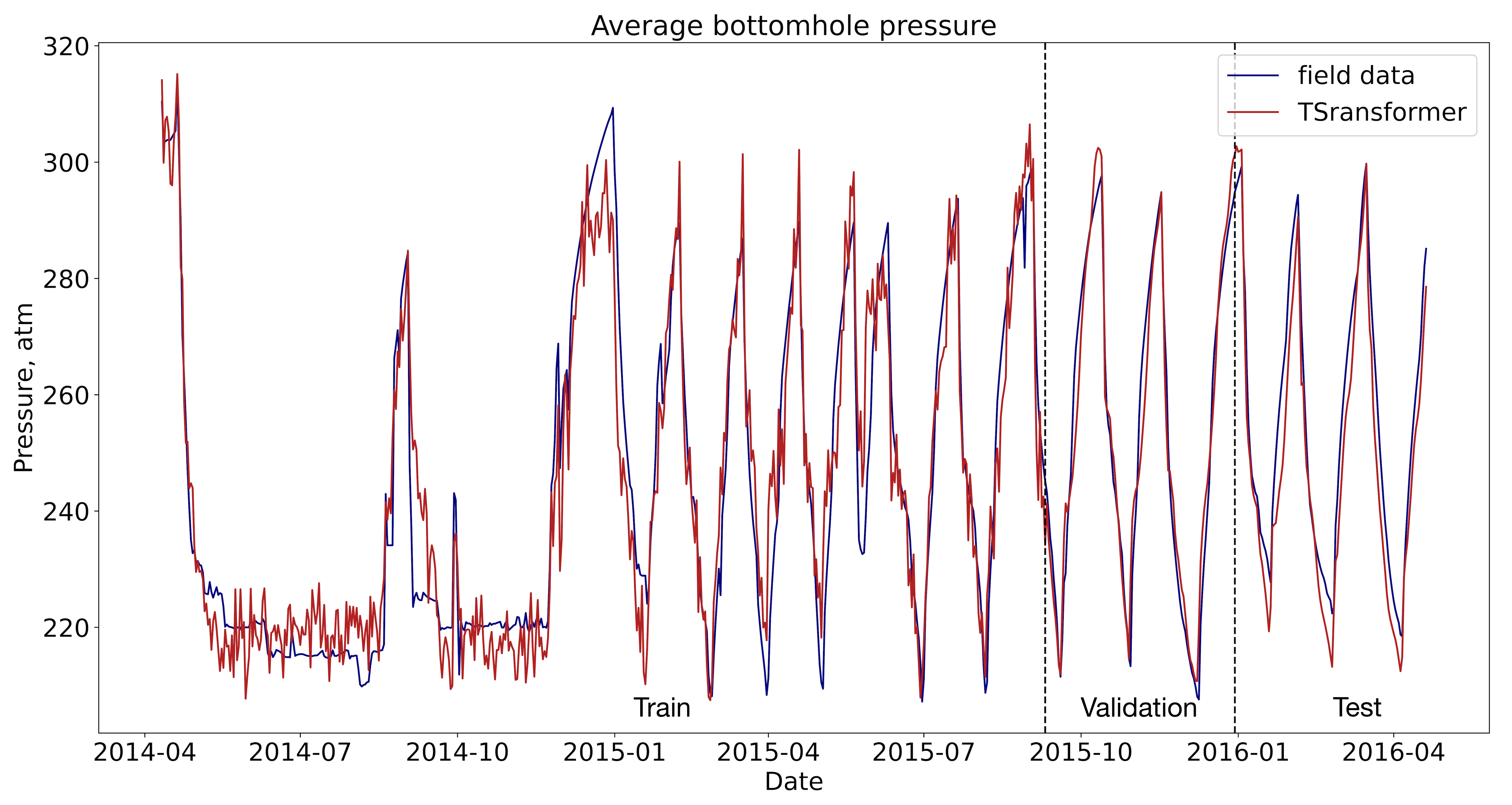}
   \caption{Prediction of the bottomhole pressure dynamics for well 15/9-F-1C by the transformer-based deep learning model.}
   \label{fig:trans_predictions}
\end{figure}

First, we compared the predictive capabilities of the recurrent neural networks and transformer for prediction of the bottomhole pressure dynamics of a single well. We trained the deep learning models on the data from each well separately. Input features include daily measurements of oil, gas, and water flow rates, bottomhole temperature, and choke size in percentage (see Appendix). The loss function is a mean squared error (MSE). The sequence length of the input samples is fourteen days ($N=14$). The best state of a surrogate model corresponds to the lowest MSE value on the validation dataset. The example transformer-based model performance on train, validation, and test sets is shown in Figure \ref{fig:trans_predictions}.

\begin{figure}[htbp]
     \centering
     \begin{subfigure}[b]{0.45\textwidth}
         \centering
         \includegraphics[width=\textwidth]{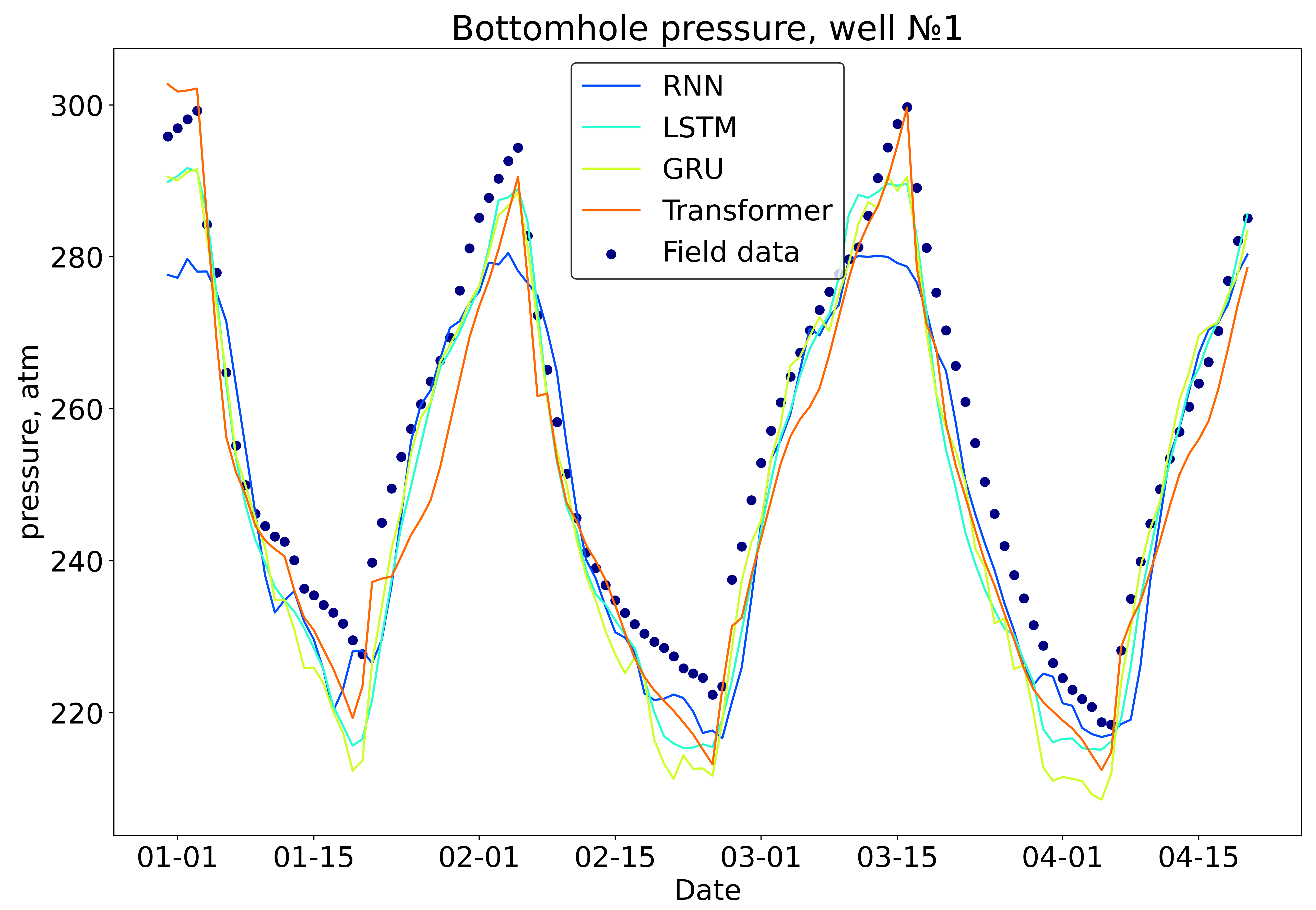}
         \caption{}
     \end{subfigure}
     \hfill
     \begin{subfigure}[b]{0.45\textwidth}
         \centering
         \includegraphics[width=\textwidth]{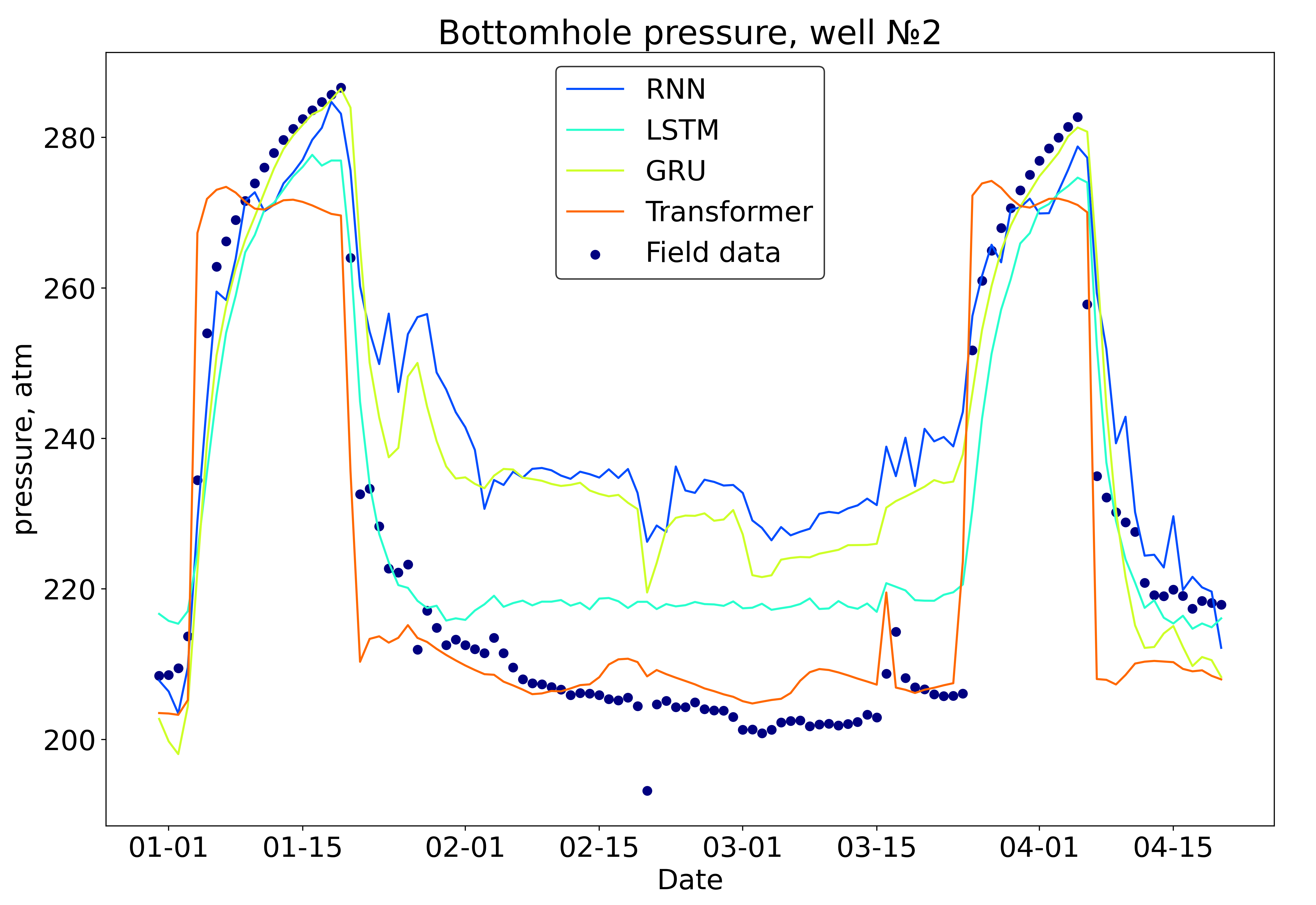}
         \caption{}
     \end{subfigure}
    \caption{Predictions of the bottomhole pressure dynamics for the test set corresponding to well 15/9-F-1C (a) and 15/9-F-15D (b) by the deep learning models based on the recurrent neural networks (vanilla RNN, LSTM, GRU) and transformer.}
    \label{fig:single_predictions}
\end{figure}

We benchmark the transformer model against the recurrent neural networks on the test set for two wells, and Figure \ref{fig:single_predictions} presents the obtained results.

In addition to the qualitative comparison of the surrogate models performance, we also carried out the quantitative one by computing the values of different metrics such as RMSE, RSE, RAE, and MAPE using the actual and estimated bottomhole pressure values according to the formulas below: 
\begin{flalign}
&\mathrm{RMSE}=\ \sqrt{\frac{\sum_{i=1}^{n}\left({\hat{y}}_i-y_i\right)^2}{n}}, \nonumber \\
&\mathrm{RSE}=\frac{\sum_{i=1}^{n}\left({\hat{y}}_i-y_i\right)^2}{\sum_{i=1}^{n}\left(\overline{y}-y_i\right)^2}, \nonumber \\
&\mathrm{RAE}=\frac{\sum_{i=1}^{n}\left|{\hat{y}}_i-y_i\right|}{\sum_{i=1}^{n}\left|\overline{y}-y_i\right|}, \nonumber \\
&\mathrm{MAPE}=\sum_{i=1}^{n}\left|\frac{{\hat{y}}_i-y_i}{y_i}\right|,
\end{flalign}
where RMSE is the root mean square error, RSE is the relative square error, RAE is the relative absolute error, MAPE is mean absolute percentage error, $y_i$ is the actual value of the target parameter at the time moment $i$, ${\hat{y}}_i$ is the predicted value of the target parameter at the time moment $i$, $\overline{y}$ is the average value of the target parameter, n is the number of datapoints. The obtained metrics are summarized in Table \ref{tab:single_metrics_well1} and Table \ref{tab:single_metrics_well2}.

\begin{table}[h]
\caption{The values of metrics calculated for the test set corresponding to well 15/9-F-1C and based on the real and predicted values of the bottomhole pressure. The best values of each metric are highlighted in red.}
\begin{center}
\label{tab:single_metrics_well1}
\begin{tabularx}{\textwidth}{ 
  | >{\centering\arraybackslash}X
  | >{\centering\arraybackslash}X 
  | >{\centering\arraybackslash}X
  | >{\centering\arraybackslash}X
  |>{\centering\arraybackslash}X|}
    \hline
    \textbf{Metric/Model} & \textbf{RNN} & \textbf{LSTM} & \textbf{GRU} & \textbf{Transformer} \\
    \hline
    RMSE & 8.39 & 7.76 & 8.12 & \textcolor{red}{7.66} \\
     \hline
    RSE & 0.13 & 0.11 & 0.12 & \textcolor{red}{0.10} \\
    \hline
    RAE & 0.34 & \textcolor{red}{0.33} & 0.34 & 0.34 \\
    \hline
    MAPE & \textcolor{red}{2.59 \%} & 2.61 \% & 2.73 \% & 2.62 \% \\
    \hline
\end{tabularx}
\end{center}
\end{table}

\begin{table}[h]
\caption{The values of metrics calculated for the test set corresponding to well 15/9-F-15D and based on the real and predicted values of the bottomhole pressure. The best values of each metric are highlighted in red.}
\begin{center}
\label{tab:single_metrics_well2}
\begin{tabularx}{\textwidth}{ 
  | >{\centering\arraybackslash}X
  | >{\centering\arraybackslash}X 
  | >{\centering\arraybackslash}X
  | >{\centering\arraybackslash}X
  |>{\centering\arraybackslash}X|}
    \hline
    \textbf{Metric/Model} & \textbf{RNN} & \textbf{LSTM} & \textbf{GRU} & \textbf{Transformer} \\
    \hline
    RMSE & 22.60 & 11.05 & 19.58 & \textcolor{red}{10.01} \\
     \hline
    RSE & 0.62 & 0.15 & 0.46 & \textcolor{red}{0.12} \\
    \hline
    RAE & 0.77 & 0.40 & 0.68 & \textcolor{red}{0.31} \\
    \hline
    MAPE & 8.92 \% & 4.42 \% & 7.89 \% & \textcolor{red}{3.24 \%} \\
    \hline
\end{tabularx}
\end{center}
\end{table}

The transformer model outperforms recurrent neural networks in bottomhole pressure prediction for both wells (Figure \ref{fig:single_predictions}, Table \ref{tab:single_metrics_well1} and Table \ref{tab:single_metrics_well2}). The self-attention mechanism of the transformer network enables to capture of the complex transient dynamics of field data. The advantage of the transformer network over RNNs is more pronounced for well 15/9-F-15D as compared to well 15/9-F-1C. The possible reason is that the test data corresponding to well 15/9-F-1C resembles the training data. Therefore, all models are close in terms of quality metrics. However, the test set related to well 15/9-F-15D is more variable. It suggests that the transformer model better reproduces true well performance behavior and captures the relationship between pressure and flow rates.

\begin{figure}[htbp]
     \centering
     \begin{subfigure}[b]{0.45\textwidth}
         \centering
         \includegraphics[width=\textwidth]{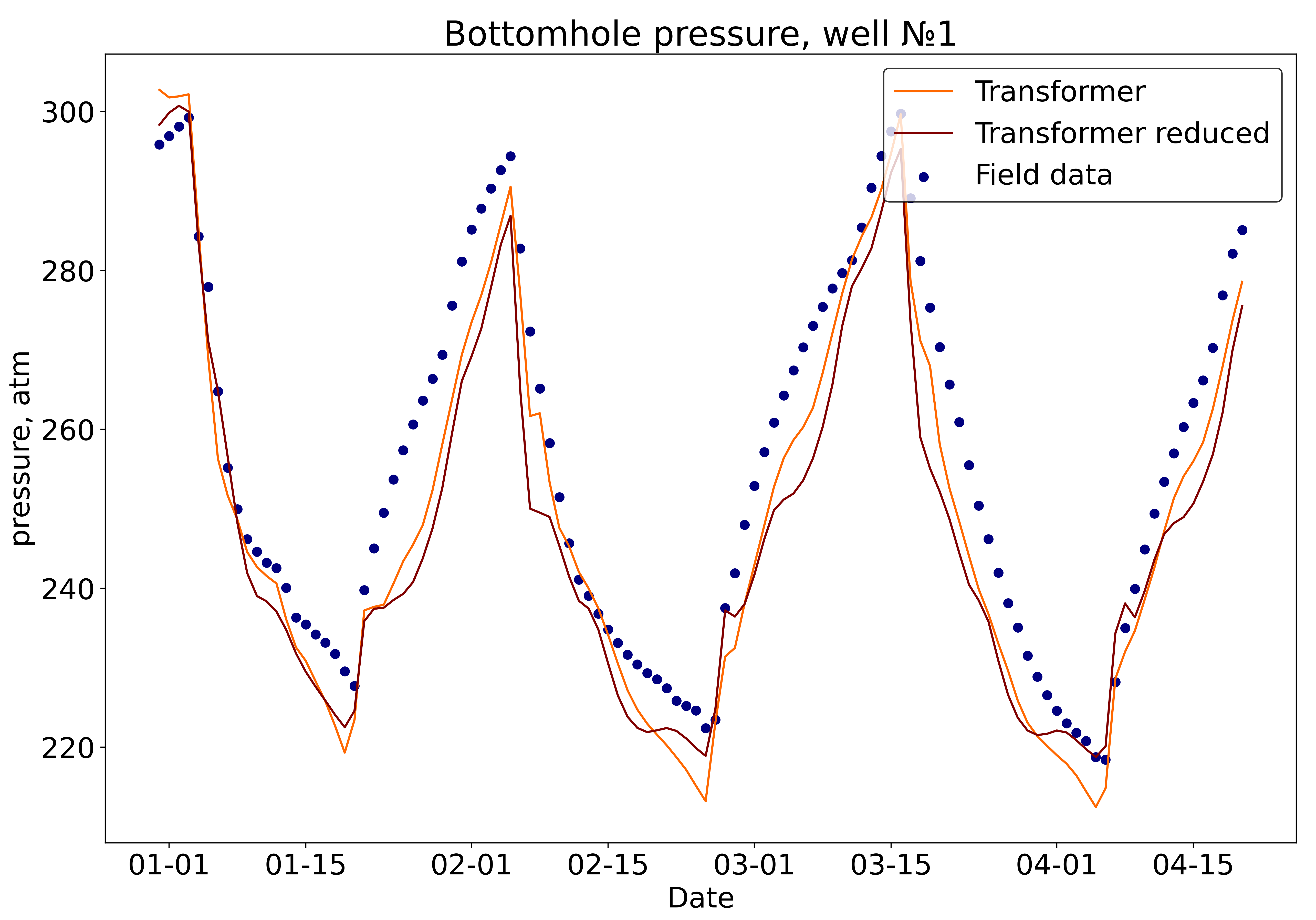}
         \caption{}
         \label{subfig:reduced_trans}
     \end{subfigure}
     \hfill
     \begin{subfigure}[b]{0.45\textwidth}
         \centering
         \includegraphics[width=\textwidth]{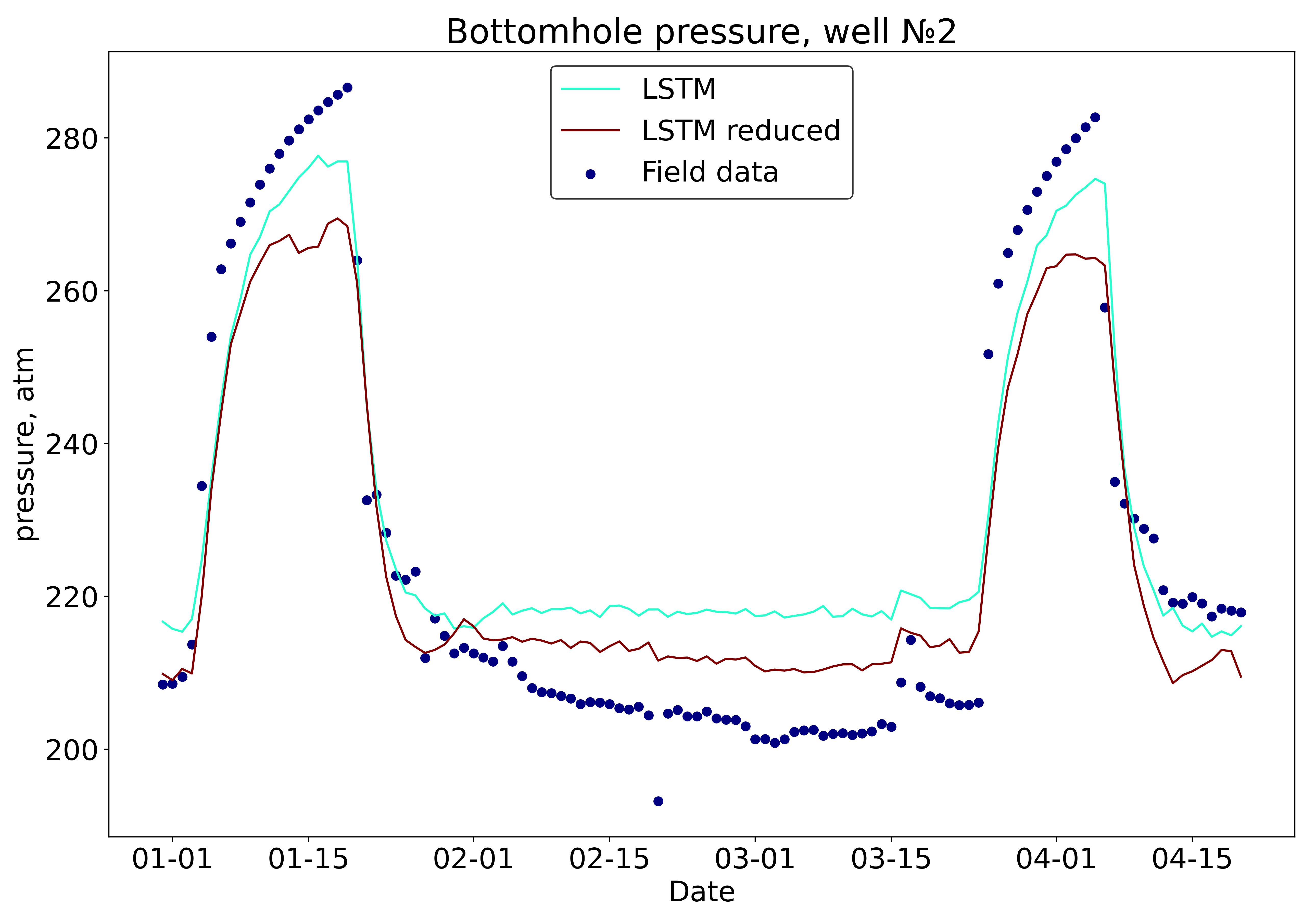}
         \caption{}
         \label{subfig:reduced_lstm}
     \end{subfigure}
    \caption{Predictions of the bottomhole pressure dynamics for the test set corresponding to well 15/9-F-1C (a) and 15/9-F-15D (b) by the deep learning models based on transformer (a) and LSTM recurrent neural network (b) trained on the full feature set and reduced feature set.}
    \label{fig:reduced_predictions}
\end{figure}

The flow rate measurement requires the availability of the multiphase flow meter, which is not always installed in a well. Moreover, the dynamics of phase flow rates can be highly oscillating and contain missing values, so that they can be unreliable features for the machine learning model. Instead of relying on flow rate data, we demonstrate that the surrogate models trained solely on choke opening size and bottomhole temperature can predict the bottomhole pressure with acceptable accuracy. Figure \ref{fig:reduced_predictions} shows the predictions of the models trained on the full and reduced feature sets for the testing datasets corresponding to wells 15/9-F-1C and 15/9-F-15D.  

We applied the transformer-based deep learning model to the data related to well 15/9-F-1C, while LSTM was used for that of well 15/9-F-15D. The transformer model trained on the reduced feature set (Figure \ref{fig:reduced_predictions}a) provides almost the same prediction quality as the model trained on the full feature set. Table \ref{tab:reduced_metrics_well1} summarizes the values of metrics computed with using the actual and estimated bottomhole pressure dynamics corresponding to the test set of well 15/9-F-1C. In turn, Figure \ref{fig:reduced_predictions}b shows the obtained results for well 15/9-F-15D, and Table \ref{tab:reduced_metrics_well2} gives the numbers of the calculated metrics. We conclude that the transformer-based surrogate model trained on the reduced feature set has lower predictive capability than the model tuned on the full feature set. However, the reversed situation is observed for the LSTM, which yields a better prediction quality after training on the reduced feature set.

\begin{table}[ht]
\caption{The values of metrics calculated for the test set corresponding to well 15/9-F-1C and based on the real and predicted values of the bottomhole pressure, which are estimated by the transformer deep learning model trained on both full and reduced feature sets.}
\begin{center}
\label{tab:reduced_metrics_well1}
\begin{tabularx}{0.8\textwidth}{ 
  | >{\centering\arraybackslash}X
  | >{\centering\arraybackslash}X 
  | >{\centering\arraybackslash}X
  | >{\centering\arraybackslash}X
  |>{\centering\arraybackslash}X|}
    \hline
    \textbf{Metric/Model} & \textbf{Transformer (full set)} & \textbf{Transformer (reduced set)}  \\
    \hline
    RMSE & 7.66 & 10.36 (+35.24 \%)   \\
     \hline
    RSE & 0.10 & 0.20 (+100 \%) \\
    \hline
    RAE & 0.34 & 0.43 (+26.47 \%) \\
    \hline
    MAPE & 2.62 \% & 3.33 \% (+27.10 \%) \\
    \hline
\end{tabularx}
\end{center}
\end{table}

\begin{table}[ht]
\caption{The values of metrics calculated for the test set corresponding to well 15/9-F-15D and based on the real and predicted values of the bottomhole pressure, which are estimated by the LSTM trained on both full and reduced feature sets.}
\begin{center}
\label{tab:reduced_metrics_well2}
\begin{tabularx}{0.8\textwidth}{ 
  | >{\centering\arraybackslash}X
  | >{\centering\arraybackslash}X 
  | >{\centering\arraybackslash}X
  | >{\centering\arraybackslash}X
  |>{\centering\arraybackslash}X|}
    \hline
    \textbf{Metric/Model} & \textbf{LSTM (full set)} & \textbf{LSTM (reduced set)}  \\
    \hline
    RMSE & 11.05 & 10.24 (-7.33 \%)   \\
     \hline
    RSE & 0.15 & 0.13 (-13.33 \%) \\
    \hline
    RAE & 0.40 & 0.37 (-7.5 \%) \\
    \hline
    MAPE & 4.42 \% & 3.87 \% (-12.44 \%) \\
    \hline
\end{tabularx}
\end{center}
\end{table}

\begin{figure}[ht]
   \centering
   \includegraphics[width=0.8\textwidth]{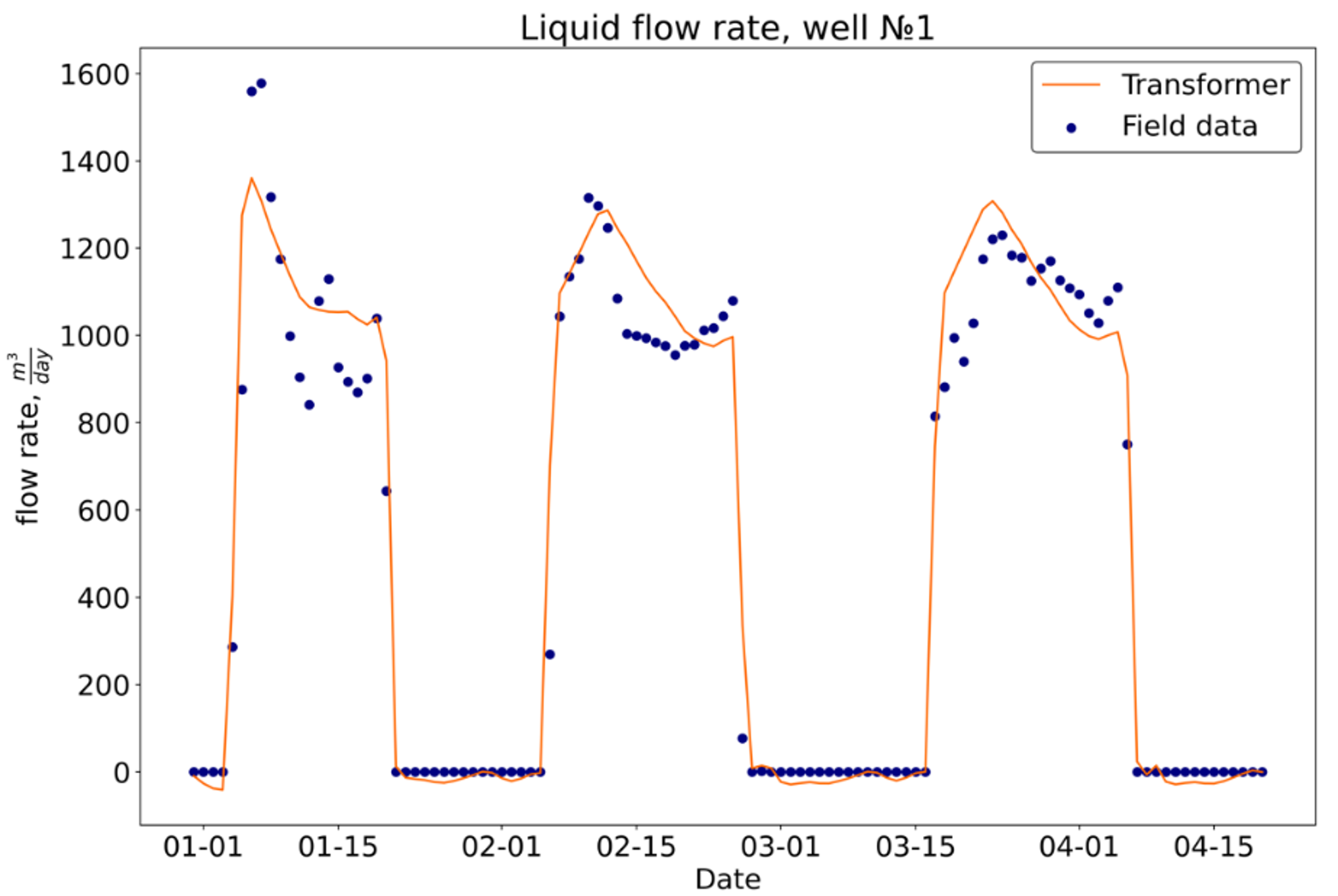}
   \caption{Prediction of the liquid flow rate dynamics for the test set corresponding to well 15/9-F-1C by the transformer-based deep learning model.}
   \label{fig:flow_rate_predictions}
\end{figure}

In addition to the model for predicting the bottomhole pressure, we also built a transformer-based model for forecasting the dynamics of the liquid flow rate. The bottomhole pressure and temperature and the choke opening size were taken as input features. We present the obtained results on the test dataset corresponding to well 15/9-F-1C in Figure \ref{fig:flow_rate_predictions}, and the computed quality metrics are written in Table \ref{tab:flow_rate_metrics}.

\begin{table}[ht]
\caption{The values of metrics calculated for the test set corresponding to well 15/9-F-1C and based on the real and predicted values of the liquid flow rate, which are estimated by the transformer-based deep learning model.}
\begin{center}
\label{tab:flow_rate_metrics}
\begin{tabularx}{0.6\textwidth}{ 
  | >{\centering\arraybackslash}X
  | >{\centering\arraybackslash}X|}
    \hline
    \textbf{Metric} & \textbf{Value} \\
    \hline
    RMSE & 107.75 \\
     \hline
    RSE & 0.04 \\
    \hline
    RAE & 0.13 \\
    \hline
\end{tabularx}
\end{center}
\end{table}

\subsection{Transfer learning technique application}
\label{sec:transfer_learning}
\begin{figure}[ht]
   \centering
   \includegraphics[width=0.8\textwidth]{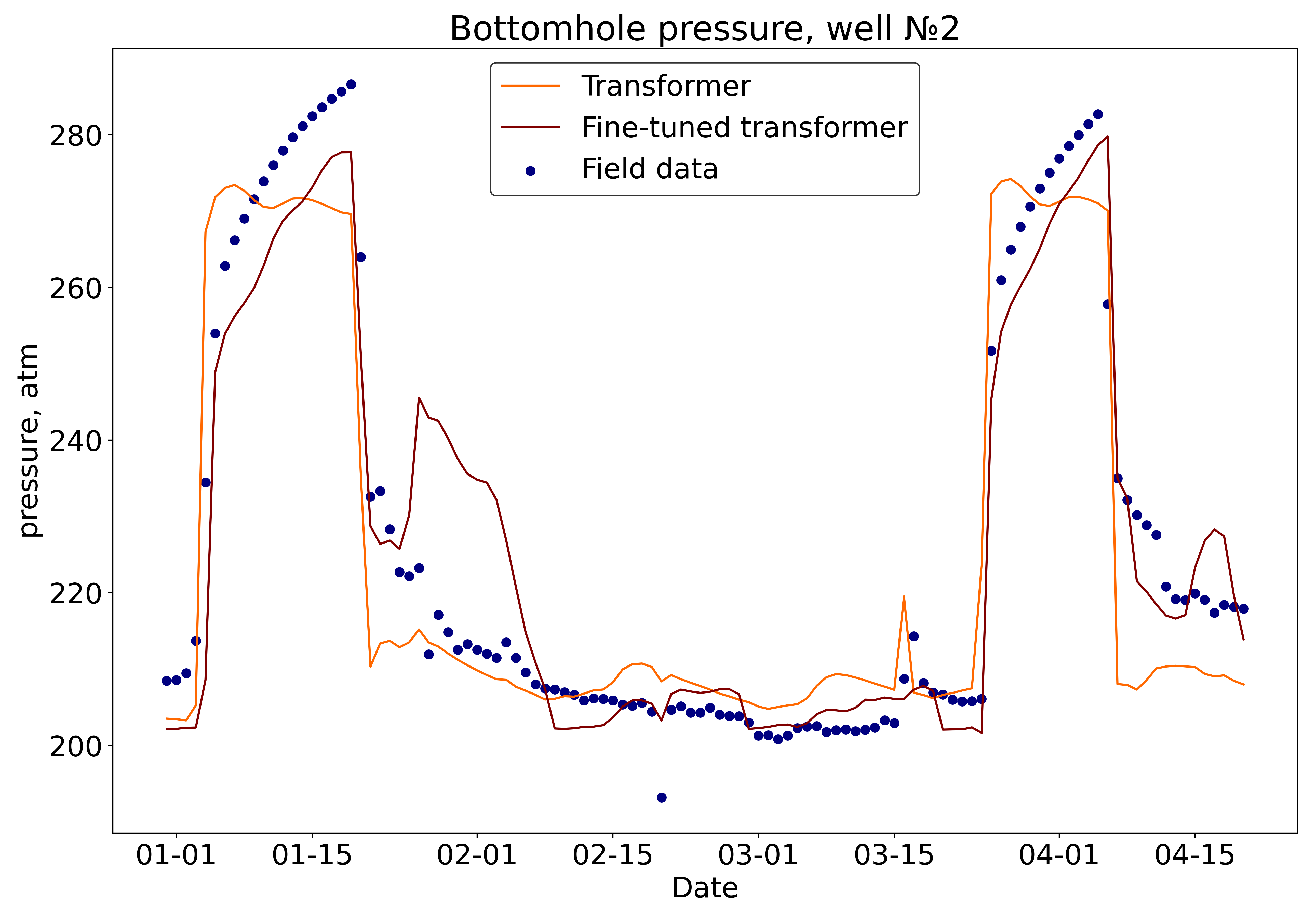}
   \caption{Predictions of the bottomhole pressure dynamics for the test set corresponding to well 15/9-F-15D by the deep learning models based on the transformer architecture trained from scratch and fine-tuned starting from the pre-trained weights on the dataset corresponding to well 15/9-F-1C.}
   \label{fig:transfer_learning}
\end{figure}

The next numerical experiment shows that the transfer learning technique that works with transformers can enable extra-quality predictions of the bottomhole pressure. We transferred weights from the model trained on the dataset corresponding to well 15/9-F-1C. We added the feed-forward network on top of the transformer and fine-tuned the whole model on the production data corresponding to well 15/9-F-15D.The resulting predictions are shown in Figure \ref{fig:transfer_learning}. 

Table \ref{tab:transfer_learning} compares the quality metrics of two transformer-based models: (i) trained from scratch and (ii) fine-tuned with pre-trained weights. The improvement is not significant since we used the data from only a single well for the pre-training stage.
\begin{table}[ht]
\caption{The values of metrics calculated for the test set corresponding to well 15/9-F-15D and based on the real and predicted values of the bottomhole pressure, which are estimated by the transformer deep learning model trained from scratch and pre-trained on the dataset from well 15/9-F-1C followed by the additional training on the dataset from the required well.}
\begin{center}
\label{tab:transfer_learning}
\begin{tabularx}{0.8\textwidth}{ 
  | >{\centering\arraybackslash}X
  | >{\centering\arraybackslash}X 
  | >{\centering\arraybackslash}X
  | >{\centering\arraybackslash}X
  |>{\centering\arraybackslash}X|}
    \hline
    \textbf{Metric/Model} & \textbf{Transformer} & \textbf{Fine-tuned pre-trained transformer}  \\
    \hline
    RMSE & 10.01 & 9.53 (-4.80 \%)   \\
     \hline
    RSE & 0.12 & 0.11 (-8.33 \%) \\
    \hline
    RAE & 0.31 & 0.28 (-9.68 \%) \\
    \hline
    MAPE & 3.24 \% & 3.00 \% (-7.41 \%) \\
    \hline
\end{tabularx}
\end{center}
\end{table}

\subsection{Global model}
\begin{figure}[ht]
     \centering
     \begin{subfigure}[b]{0.45\textwidth}
         \centering
         \includegraphics[width=\textwidth]{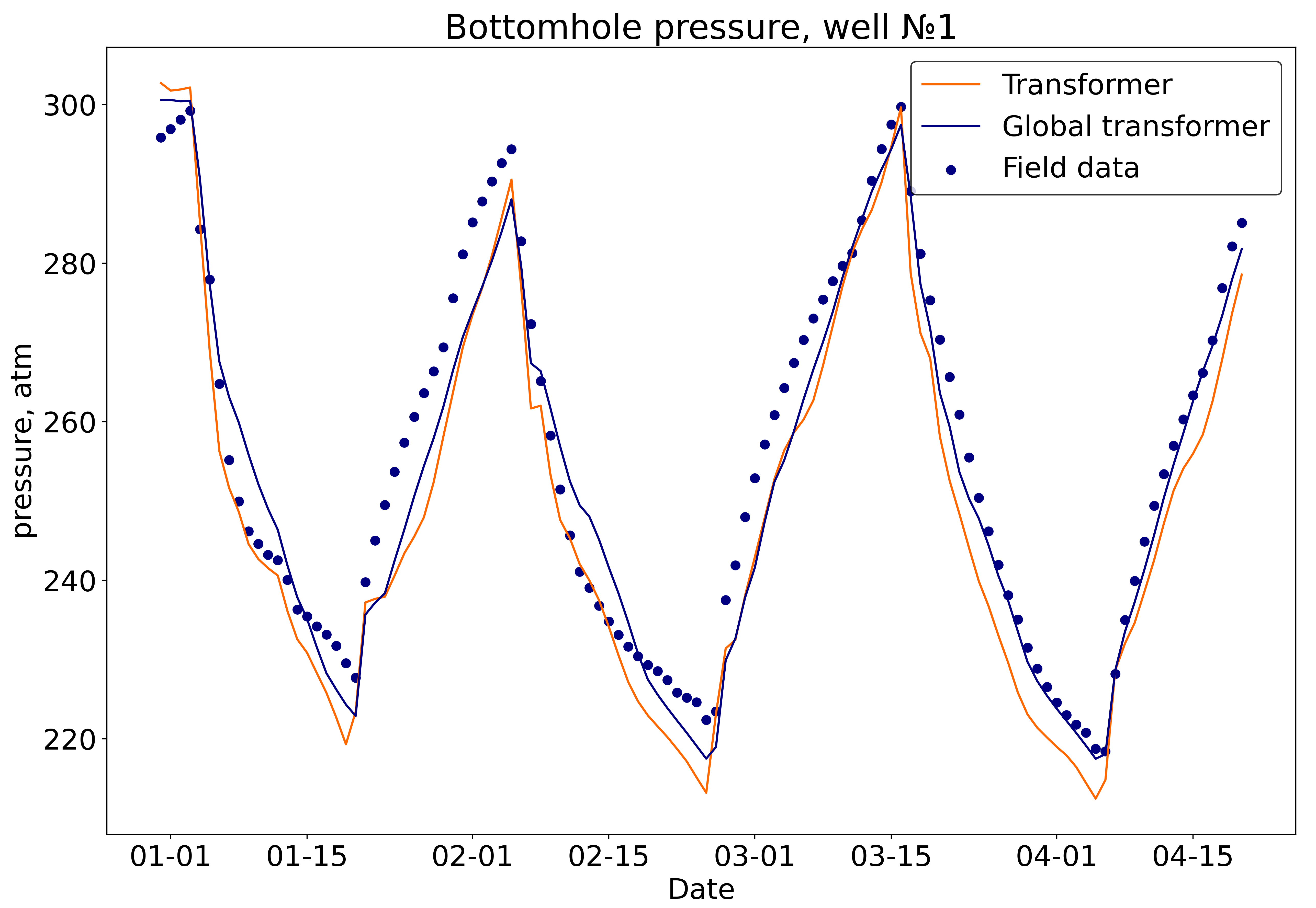}
         \caption{}
     \end{subfigure}
     \hfill
     \begin{subfigure}[b]{0.45\textwidth}
         \centering
         \includegraphics[width=\textwidth]{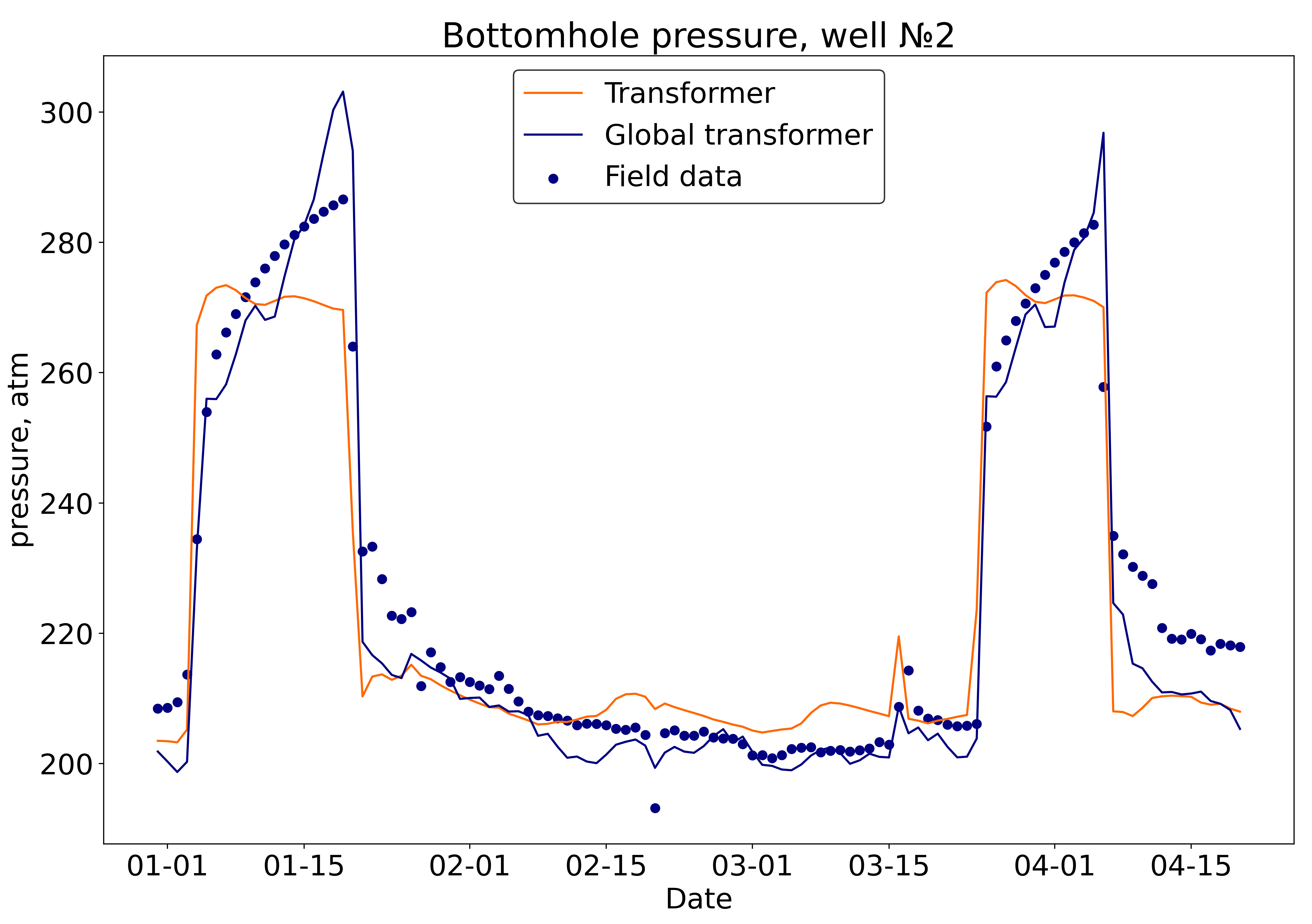}
         \caption{}
     \end{subfigure}
    \caption{Predictions of the bottomhole pressure dynamics for the test set corresponding to well 15/9-F-1C (a) and 15/9-F-15D (b) by the single-well and global deep learning models based on transformer.}
    \label{fig:global_model}
\end{figure}

The final numerical experiment is devoted to the global transformer model. We analyzed the data from the same wells (15/9-F-1C and 15/9-F-15D) and combined them into a single dataset. The predictions of the global model are demonstrated in Figure \ref{fig:global_model} for each considered well. Moreover, we add the forecasts via the single-well model for comparison purposes. The quality metrics values with the relative deviation from the numbers corresponding to the single-well model performance are summarized in Table \ref{tab:global_metrics}.

\begin{table}[ht]
\caption{The values of metrics calculated for the test sets for wells 15/9-F-1C and 15/9-F-15D and based on the real and predicted values of the bottomhole pressure, which are estimated by the global model relied on the transformer architecture. In brackets, we write the relative difference between the values of metrics obtained via the global and single-well models.}
\begin{center}
\label{tab:global_metrics}
\begin{tabularx}{\textwidth}{ 
  | >{\centering\arraybackslash}X
  | >{\centering\arraybackslash}X 
  | >{\centering\arraybackslash}X
  | >{\centering\arraybackslash}X
  |>{\centering\arraybackslash}X|}
    \hline
    \textbf{Well/Metric} & \textbf{RMSE} & \textbf{RSE} & \textbf{RAE} & \textbf{MAPE} \\
     \hline
     15/9-F-1C & 5.76 (-24.80 \%)  & 0.06 (-40 \%) & 0.24 (-29.41 \%) & 1.85 \% (-29.39 \%) \\
    \hline
     15/9-F-15D  & 7.97 (-20.38 \%) & 0.08 (-33.33 \%) & 0.23 (-25.81 \%) & 2.39 \% (-26.23 \%) \\
\hline
\end{tabularx}
\end{center}
\end{table}

By comparing the RMSE values, we conclude that the global model outperforms the single-well model for the well 15/9-F-1C by 25\% and the model for well 15/9-F-15D by 20\%. If the data from several or even many wells working at the same oil field is available, it is possible to train the global model to forecast the bottomhole pressure or liquid flow rate, accounting for the effect of the influence of wells on each other (interference) during the production process.

\section{Conclusions}
We have developed a novel approach to simulate the transient production of an oil well based on the deep learning algorithm called transformer. We have compared the predictive capability of the transformer with the performance of the RNN-based models in the prediction of the bottomhole pressure dynamics of a single well. The self-attention mechanism of the transformer model can learn the complex transient dependencies in field data between the target parameter (for example, bottomhole pressure) and input features, which are the measured parameters of a well (for example, flow rates, bottomhole temperature, and choke opening size) improving the accuracy of predictions of the target parameter as compared to the RNNs. Moreover, the transformer can process the entire input sequences of various lengths; in contrast, the number of cells in recurrent networks is often limited. We have applied the transfer learning technique, which enables the transformer to enhance the forecasting ability to predict the target property for one well with the initial tuning of the model weights on the data related to the other well. The transfer learning approach would benefit from the increase in the number of wells used during the pre-training stage. The global model trained on the data from two wells allows for improving the quality of the predictions significantly. 

The possible continuation of the work will include the transfer learning application to the global transformer model, further investigation of the transformer-based model ability to predict flow rate and temperature as well as to use the transformer in well-testing interpretation to avoid expensive shut-ins.

\clearpage
\newpage

\section*{Appendix}

\begin{figure}[htbp]
  \centering
  \includegraphics[width = \textwidth, height=0.85\textheight, keepaspectratio]{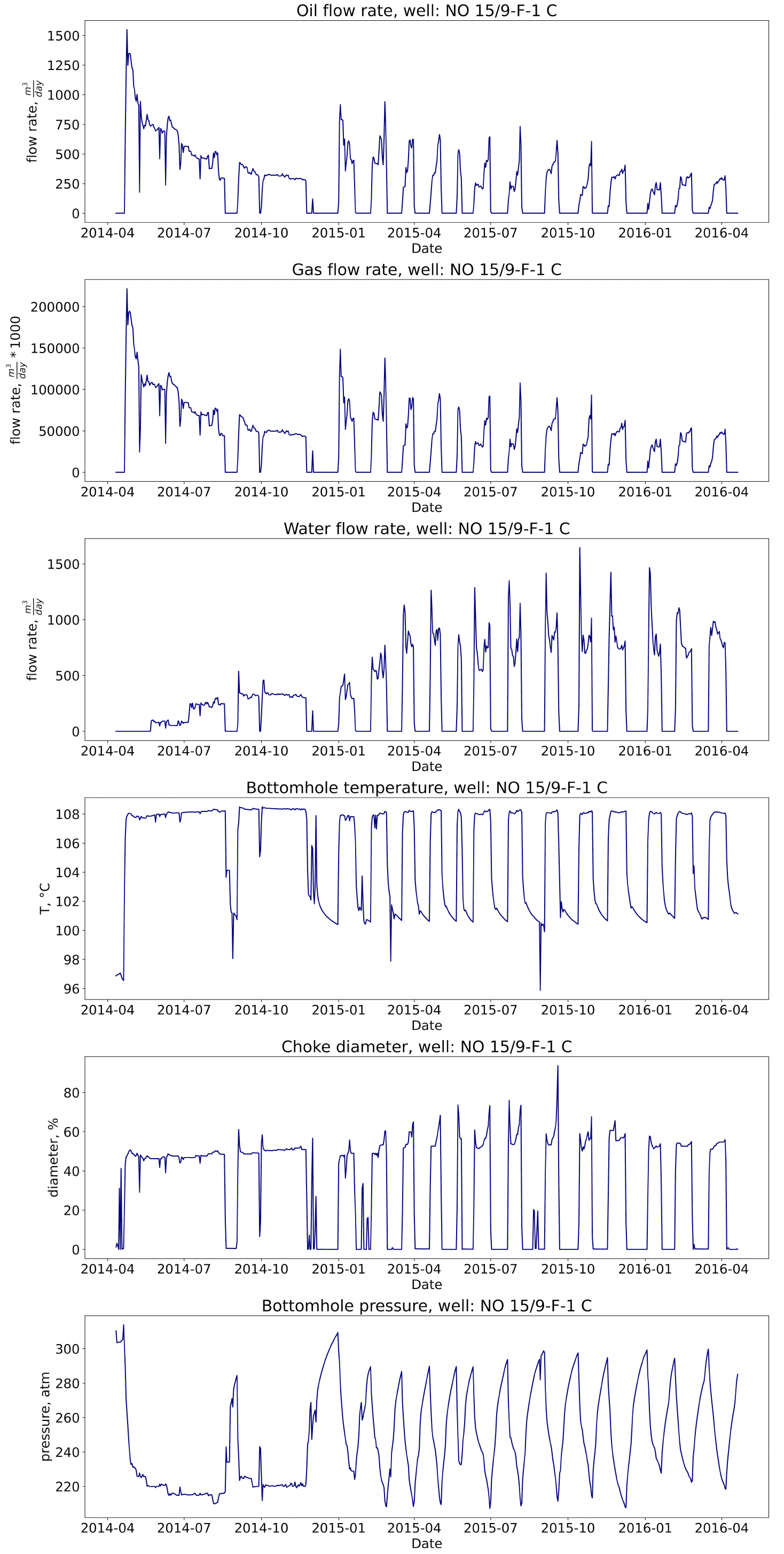}
  \caption*{Field data – oil well 15/9-F-1C.}
  \label{fig:1C_features}
\end{figure}

\begin{figure}[htbp]
  \centering
  \includegraphics[width = \textwidth, height=0.85\textheight, keepaspectratio]{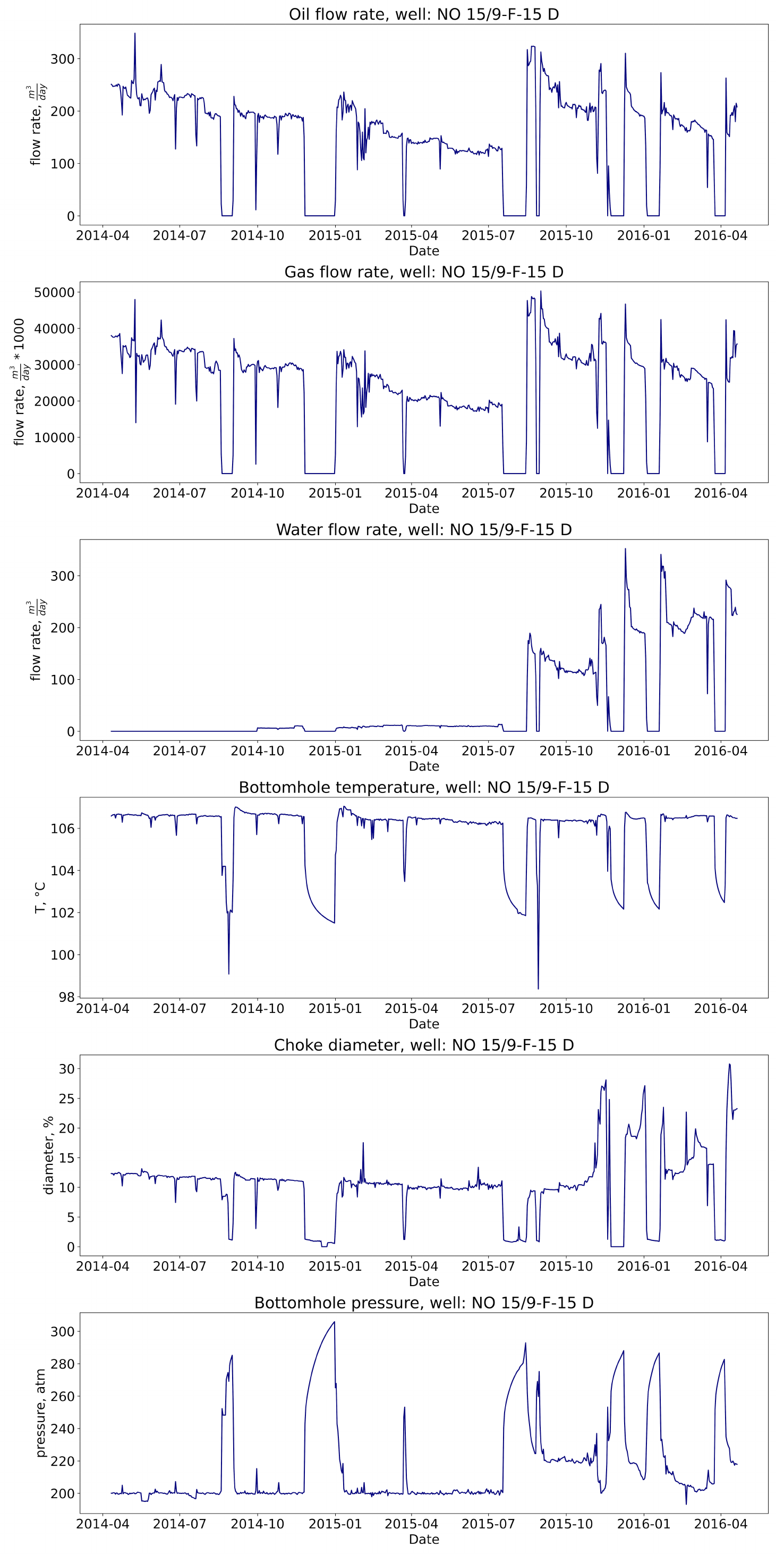}
  \caption*{Field data – oil well 15/9-F-15D.}
  \label{fig:15D_features}
\end{figure}


\clearpage
\newpage

\bibliographystyle{unsrt}  
\bibliography{references}  







\end{document}